\documentclass[review]{elsarticle}

\usepackage{times}
\usepackage{epsfig}
\usepackage{graphicx}
\usepackage{amsmath}
\usepackage{amssymb}
\usepackage{multirow}
\usepackage{color}
\usepackage{eso-pic}
\usepackage{xspace}
\usepackage{comment}
\usepackage{booktabs}

\makeatletter
\DeclareRobustCommand\onedot{\futurelet\@let@token\@onedot}
\def\@onedot{\ifx\@let@token.\else.\null\fi\xspace}

\def\ie{\emph{i.e}\onedot}

\def\etal{\emph{et al}\onedot}
\makeatother

\usepackage{amssymb}
\usepackage{lineno,hyperref}
\journal{Journal of \LaTeX\ Templates}

\begin{document}

\begin{frontmatter}

%% Title, authors and addresses

%% use the tnoteref command within \title for footnotes;
%% use the tnotetext command for theassociated footnote;
%% use the fnref command within \author or \address for footnotes;
%% use the fntext command for theassociated footnote;
%% use the corref command within \author for corresponding author footnotes;
%% use the cortext command for theassociated footnote;
%% use the ead command for the email address,
%% and the form \ead[url] for the home page:
%% \title{Title\tnoteref{label1}}
%% \tnotetext[label1]{}
%% \author{Name\corref{cor1}\fnref{label2}}
%% \ead{email address}
%% \ead[url]{home page}
%% \fntext[label2]{}
%% \cortext[cor1]{}
%% \address{Address\fnref{label3}}
%% \fntext[label3]{}

\title{An Adversarial Human Pose Estimation Network Injected with Graph Structure}

%% use optional labels to link authors explicitly to addresses:
%% \author[label1,label2]{}
%% \address[label1]{}
%% \address[label2]{}

\author[NWPU,NWPUE]{Lei Tian\fnref{correspondingauthor}}
%\ead{tianlei_nwpu@mail.nwpu.edu.cn}

\author[NWPU,NWPUE]{Peng Wang\fnref{correspondingauthor}}
%\cortext[correspondingauthor]{The first two author contribute equally to this paper.}
%\ead{peng.wang@nwpu.edu.cn}
\fntext[correspondingauthor]{The first two authors contribute equally to this paper.}

\author[NWPU,NWPUE]{Guoqiang Liang\fnref{mycorrespondingauthor}}
\fntext[mycorrespondingauthor]{Corresponding author: Guoqiang Liang}
%\cortext[mycorrespondingauthor]{Corresponding author: Guoqiang Liang}
\ead{gqliang@nwpu.edu.cn}

\author[Adelaide]{Chunhua Shen}
%\ead{chunhua.shen@adelaide.edu.au}

% full name
\address[NWPU]{School of Computer Science, Northwestern Polytechnical University, Xi'an, China}
\address[NWPUE]{National Engineering Laboratory for Integrated Aero-Space-Ground-Ocean Big Data Application Technology, China}
\address[Adelaide]{School of Computer Science, The University of Adelaide, Australia}

\begin{abstract}
%% Text of abstract 

Because of the invisible human keypoints in images caused by illumination, occlusion and overlap, it is likely to produce unreasonable human pose prediction for most of the current human pose estimation methods. 
In this paper, we design a novel generative adversarial network (GAN) to improve the localization accuracy of visible joints when some joints are invisible.
The network consists of two simple but efficient modules, \ie, Cascade Feature Network (CFN) and Graph Structure Network (GSN).
First, the CFN utilizes the prediction maps from the previous stages to guide the prediction maps in the next stage to produce accurate human pose.
Second, the GSN is designed to contribute to the localization of invisible joints by passing message among different joints.
According to GAN, if the prediction pose produced by the generator $G$ cannot be distinguished by the discriminator $D$, the generator network $G$ has successfully obtained the underlying dependence of human joints. 
We conduct experiments on three widely used human pose estimation benchmark datasets, \ie, LSP, MPII and COCO, whose results show the effectiveness of our proposed framework.

\end{abstract}

\begin{keyword}
%% keywords here, in the form: keyword \sep keyword
Human pose estimation \sep Cascade feature network \sep Graph structure network \sep Generative adversarial network

%% PACS codes here, in the form: \PACS code \sep code

%% MSC codes here, in the form: \MSC code \sep code
%% or \MSC[2008] code \sep code (2000 is the default)

\end{keyword}

\end{frontmatter}

%\linenumbers

%% main text

\section{Introduction}
\label{}

Human pose estimation refers to predict the specific location of human keypoints from an image. 
It is a fundamental yet challenging task for many computer vision applications like intelligent video surveillance and human-computer interaction.
However, due to illumination, occlusion and flexible body poses, human pose estimation is still difficult, especially for locating invisible keypoints.

In recent years, due to the development of deep convolutional neural networks (CNNs)~\cite{xiao2018simple,zhu2019rotated}, there has been significant progress on the problem of human pose estimation.
Newly proposed methods mainly use CNNs to predict the heatmaps of body joints. 
However, it will be very tough for purely CNN-based methods to regress accurate heatmaps in the face of severely occluded body parts, which are caused by surrounding people or backgrounds that are very similar to body parts.

In contrast, people can always get accurate results for occluded joints by resorting to the local features and surrounding information in the image. For example, after getting the position of a knee joint, we can infer the position of corresponding ankle joint on the basis of the connection between knees and ankles, and vice versa. Moreover, even for extremely occluded body, we can distinguish reasonable and unreasonable poses and infer potential poses.
During the process of inferring  the location of invisible joints, the dependence correlation of human body joints has played an important role, which is also pointed out in~\cite{zhang2019human}.
However, the complexity of pose estimation methods will become much higher if we directly add a new module to capture joints dependence.

Considering recent success of Generative Adversarial Networks (GANs)~\cite{goodfellow2014generative}, we propose a GAN based human pose estimation framework to explore the dependence relationship among different joints while maintaining the original inference complexity. 
Specifically, since the internal connection of body joints can be regarded as a graph structure, we design a new graph structure network based on the graph neural network. The pose heatmaps from the generator $G$ are used as the input of pose discriminator $D$, whose goal is to determine whether the input pose is geometrically reasonable. In learning, the generator based CFN is trained to deceive the discriminator that its results are all reasonable. In other words, the generator is guided to learn the dependence correlation of the human body joints integrated in $D$ network. Once converged, the generator $G$ will successfully learn the internal dependence of body joints. To evaluate the proposed approach, we conduct experiments on three public datasets for human pose estimation, \ie, LSP, MPII and COCO. The experimental results show that our method has obtained more plausible pose predictions compared to previous methods.

In a word, our main contributions can be summarized as follows:

\noindent{\bf (i)} 
We propose to use graph structure to model the relationship of body joints. Specifically, we regard the nodes in graph structure as human joints representation and the edges as the dependence relations of human joints. Through message pass among nodes, the complex relations of joints can be captured.

\noindent{\bf (ii)} 
Our model is based on adversarial learning and can be trained end-to-end. Since the discriminator is not needed during testing, the complexity of pose estimation does not increase.
Experiments on three public datasets show the effectiveness of our method.

\section{Related Work}\label{}
This section will review some related works on human pose estimation, graph structure network and generative adversarial network.
\subsection{Human Pose Estimation} 
Along with the introduction of CNN, human pose estimation based on CNN~\cite{liang2017limb,wang2019deep} has attracted an increasing attention. The mainstream methods for body joints detection can be divided into two kinds. One is directly regressing joints' locations~\cite{fan2015combining,carreira2016human} while the other is producing joints heatmaps, where the location with the maximum value is selected as the predicted results~\cite{tompson2015efficient,sun2019deep}. 
DeepPose, one of the earliest CNN-based method proposed by Toshev \etal~\cite{toshev2014deeppose}, regresses location of joints with a two-steps CNN. In order to further improve performance, Fan \etal~\cite{fan2015combining} explore the relationship between global features and local features. To encompass both input and output spaces, Carreira \etal~\cite{carreira2016human} iteratively refine prediction results by concatenating the image with the body joint predictions from the previous steps. Tompson \etal~\cite{tompson2015efficient} generate heatmaps by using the multiple resolution convolutional networks to fuse the features, and employ the post-processing by using Markov Random Field (MRF). Due to its excellent performance, hourglass Network~\cite{newell2016stacked} has become a main framework, which is a bottom-up and top-down architecture with residual blocks. 
SimpleBaseline Network~\cite{xiao2018simple} estimates heatmaps by integrating a few deconvolutional layers on a backbone network in a much simpler way. 
Carissimi \etal~\cite{carissimi2018filling} consider missing joints as noise in the input and use an autoencoder-based solution to enhance the pose prediction.
Zhao \etal~\cite{zhao2020perceiving} design an occlusion aware network to predict both heatmaps and visible values of joints simultaneously, so as to grasp the degree of occlusion of a pose and set a much fairer score. 

Like the SimpleBaseline network, our generator $G$ is also a fully convolutional network with ``conv-deconv'' architecture. However, our network integrates the multi-stage features of the backbone network into the deconvolutional layers through skip layers connection. Here, we use the ResNet as our backbone. However, our method is not restricted to the ResNet. And newly proposed network architecture can be easily used to further improve the performance.
%~\cite{he2016deep}

\subsection{Graph Structure Network} 
The dependence correlation of human joints is very important in human pose estimation. In most of traditional methods for human pose estimation, graphical model, especially a tree-structure model is used to capture the relationship among human joints~\cite{yang2012articulated}. Recently, this graphical model is combined with convolution neural networks. Chu \etal~\cite{chu2016structured} propose a structured feature learning model to capture the relation of joints on feature maps. 
Zhao \etal~\cite{zhao2019semantic} design a new method which operates on regression tasks with graph-structured data to predict 3D human pose from 2D joints as well as image features.
Liu \etal~\cite{liu2019feature} design the Graphical ConvLSTM to enable the intermediate convolutional feature maps to perceive the graphical long short-term dependency among different body parts.

To process the graph-structured data, Graph Neural Network (GNN) is first proposed in~\cite{scarselli2008graph}. Different from CNN, which is suitable for the data in the Euclidean space, GNN mainly deals with the data from non-Euclidean domains. 
Current GNN methods generally follow two directions. One direction is the spectral domain~\cite{defferrard2016convolutional,levie2018cayleynets}, where the neural networks are applied to the whole graph. 
The other line is spatial domain~\cite{niepert2016learning,zhang2019human}, where the neural networks are applied recurrently on each node of the graph. The state of each node is updated according to its previous state and information aggregating from neighboring nodes.
Recently, the Gated Graph Neural Network (GGNN)~\cite{li2015gated} has been widely used in some computer vision tasks. The GGNN adopts Gated Recurrent Units~\cite{cho2014learning} and unrolls the recurrence for a fixed number of steps $T$ to update the hidden state of the nodes. The final GGNN computes gradients by using the backpropagation through temporal algorithm. Compared with~\cite{zhang2019human}, our work is based on GAN and the complexity of inference does not increase.

\subsection{Generative Adversarial Network} 
Generative Adversarial Networks(GANs)~\cite{goodfellow2014generative} have been widely studied in many computer vision tasks, such as image super-resolution~\cite{zhao2019simultaneous}, image inpainting~\cite{jiao2019multi}, style transfer~\cite{fang2020identity}, and human pose estimation~\cite{wandt2019repnet,peng2018jointly}. 
For example, RepNet~\cite{wandt2019repnet} uses an adversarial training method to learn a mapping from a distribution of 2D poses to a distribution of 3D poses.
Peng \etal~\cite{peng2018jointly} apply adversarial network to human pose estimation to jointly optimize data augmentation and network training.
Human pose estimation can been regarded as a image-to-heatmap translation problem, which can be well implemented by our proposed framework. 

Different from previous GANs, we use the internal structure of human joints to construct a graph and incorporate GGNN into the discriminator model to distinguish the reasonable poses from unreasonable prediction results. 

\begin{figure}[t!]
\begin{center}
\includegraphics[width=0.96\linewidth]{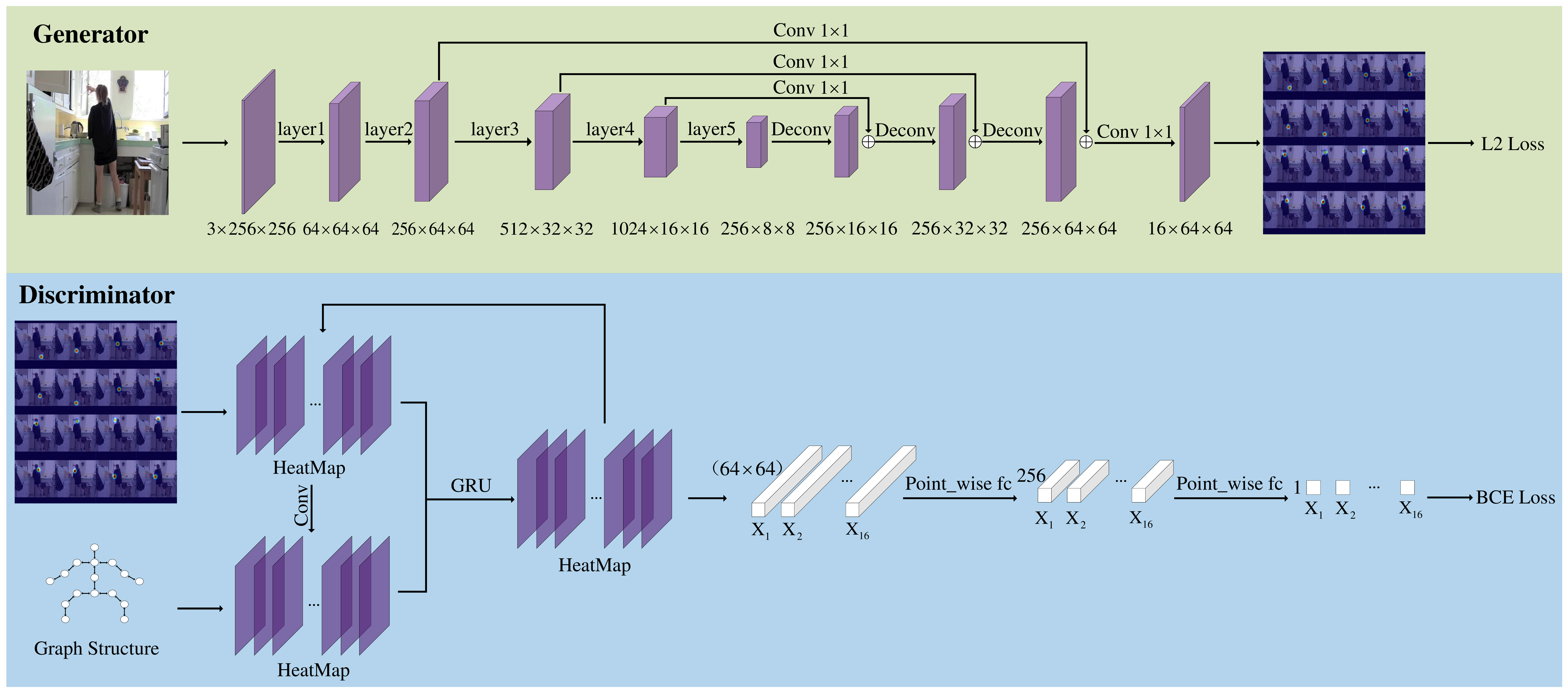}
\end{center}
\vspace{-7mm}
\caption{The overall framework of proposed method. The upper part is the generator with CFN and the lower part is the discriminator with GSN.}
\label{fig:framework}
\end{figure}

\section{Our Method}\label{}
Our human pose estimation framework consists of two parts: a generator and a discriminator, which are shown in upper part and bottom part of Figure~\ref{fig:framework} respectively. Our generator is a cascade feature network, which aims at regressing pose heatmaps in a skip layer connection way. Specifically, it takes a RGB image as input and produce a heatmap for each joint. Each value in a heatmap of a joint denotes the likelihood that the corresponding joint locates at that coordinate. Therefore, we take the coordinate corresponding to the maximum value in a heatmap as the final location of this joint. The size of predicted heatmaps is $N \times H \times W$, where $N$, $H$, $W$ represent respectively the channel size, the height and the width. The specific value of $N$ depends on the dataset. For MPII and LSP, $N$ is equal to $16$ and $14$, respectively. And due to the pooling, the height and width of heatmaps is a quarter of the original image size. In other words, if the size of images is $256 \times 256$, $H$ and $W$ will be 64. 

Generally, the generator network can be trained like a normal deep convolution network. However, it is possible to generate human pose estimations with very low confidence or even wrong estimation results. Therefore, we introduce the discriminator network $D$ into the whole framework to make the generator network correct poor estimations. To capture the internal structure dependence of human body, we design a graph neural network based on the gated graph neural network. Its inputs are heatmaps generated by the generator network or ground-truth heatmaps. The discriminator will produce a $N$-dimension vector in the range of $[$0$,$1$]$, whose value represents the joint localization quality of an input. Through adversarial learning between $G$ and $D$, the generator network $G$ will learn the dependence of human joints and output better pose prediction.

\subsection{Cascade Feature Network (CFN)}
\label{word:CFN}
Inspired by the successful idea in the previous human pose estimation~\cite{chen2018cascaded,xiao2018simple}, we also integrate cascade features through an encoding-decoding process. In the encoding process, features generated by different layers have different semantic levels. As described in~\cite{jiao2019multi}, lower level features focus on local semantic information while the upper level features focus on global semantic information. If we integrate these heatmaps to the decoding process, it will help to distinguish different body joints so as to improve the accuracy of final prediction. 

In order to effectively combine features at different semantic levels, we utilize a cascade feature network (CFN) shown in the upper part of Figure~\ref{fig:framework}. Specifically, a $1\times1$ convolution layer with the step size of 1 and without padding is used to change the channel of coarse prediction feature maps generated in the encoding stage. Then, the resulted features are combined with the corresponding image features in the decoding stage by element-wise addition. The combined feature maps are sent to the next stage of decoding. The final predicted results are obtained through the last $1 \times 1$ convolution layer, which is used as a linear classifier to map features to heatmaps.

Considering that the visibility of each human joint in images is different, we employ the following loss function for CFN:
\begin{equation}
\label{eq:mse_loss}
L_{\mbox{G}} = \frac{1}{N}\sum_{n=1}^{N}v_{n}\big \| X_{n}-Y_{n} \big \|,
\end{equation}
where $\big \| \cdot  \big \|$ represents the Euclidean distance. $N$, $v_{n}\in {\big \{ 0,1 \big \}}$ represent respectively the number of joints and the visibility of the $n$-th pose joint. $X_{n}\in \mathbb{R}^{64\times 64}$, $Y_{n}\in \mathbb{R}^{64\times 64}$ represent respectively the predicted heatmap and the ground-truth heatmap for the $n$-th joint.

\subsection{Graph Neural Network (GNN)}\label{word:GNN}
Before detailing the specific architecture of discriminator, we first introduce graph neural network (GNN), which is the most important module in our discriminator. GNN is more and more used in computer vision, especially in processing the data with graph structure. Obviously, the human keyjoints can be regarded as a natural graph structure. Therefore, we adopt a graph structure to effectively integrate the internal dependence of pose joints.

Here, we introduce our constructed GNN for human pose estimation. The input of GNN is a graph, which can be expressed as $Graph = \big \{V, E\}$, where $V$ and $E$ represent the set of nodes and edges in a graph respectively. Each node $v\in V$ represents a body joint while an edge represents the relationship between two nodes. Moreover, a node is associated with a hidden state vector $i_{n}$, which is gradually updated in the GNN. When updating the hidden state vector of node $v$ at time step $t$, the input is the combination of its current state vector $i_{n}^{t}$ and the information $j_{n}^{t}$ passed from its neighboring nodes $M_{n}$. The overall process of GNN can be expressed as:
\begin{align}
\label{eq:lm_loss}
j_{n}^{t} &= {\mbox{F}}(i_{m}^{t-1}\mid m\in M_{n}), \\
i_{n}^{t} &= {\mbox{H}}(i_{n}^{t-1},j_{n}^{t}),
\end{align}
where $F$ and $H$ represent respectively the function to collect information from neighboring nodes and to update hidden state information of a node.

\noindent{\bf Graph Construction.} 
The node connection of a graph determines information of which nodes will be collected for a specific node. Intuitively, it is simple and convenient to construct the graph for pose estimation in the way of full connection to pass information among nodes with one step, but this will destroy the internal spatial connection of body joints. Therefore, we build a tree-based graph structure to retain the relationship of neighboring joints as shown in Figure~\ref{fig:framework}. And note that the information passing in this graph is bidirectional.

\noindent{\bf Graph Propagation.} 
Before learning the current hidden state of each node, it needs to obtain the information passed from the neighboring nodes by using the function $F$. We formulate the collecting process as following:
\begin{equation}
\label{eq:lm_loss2}
j_{n}^{t}=\sum _{n^{'}\in \Omega_n } W_{n, n^{'}}i_{n}^{t-1},
\end{equation}
where $\Omega_n$ is a set of all neighboring nodes of the $n$-th node, $W_{n, n^{'}}$ is the convolution weight of the edge between nodes $n$ and $n^{'}$.

Next we need to update the state vector. In this paper, we adopt the gated graph neural network with GRU~\cite{li2015gated} to fully explore the dependence information of body joints. Specifically, after aggregating the information from neighboring nodes of the $n$-th node, the hidden state of $n$-th node is updated through GRU. The whole process can be represented as follows:
\begin{equation}
\begin{aligned}
\label{eq:lm_loss3}
z_{n}^{t} &=\sigma (W_{n}^{z}j_{n}^{t} + U_{n}^{z}i_{n}^{t-1}), \\
r_{n}^{t} &=\sigma (W_{n}^{r}j_{n}^{t} + U_{n}^{r}i_{n}^{t-1}), \\
\tilde{h}_{n}^{t} &=tanh(W_{n}^{h}j_{n}^{t} + U_{n}^{h}(r_{n}^{t}\odot  i_{n}^{t-1})), \\
i_{n}^{t} &=(1-z_{n}^{t})\odot i_{n}^{t-1} + z_{n}^{t}\odot \tilde{h}_{n}^{t},
\end{aligned}
\end{equation}
where $W_{n}^{z}, W_{n}^{r}, W_{n}^{h}, U_{n}^{z}, U_{n}^{r}, U_{n}^{h}$ are the convolution weights to be learned in the graph network, $\sigma$ denotes the $sigmoid$ function.

\subsection{Discriminator Network}\label{}
When some body parts are seriously occluded, it will be difficult to produce a reasonable prediction only through the network in the section~\ref{word:CFN}. However, human can get more accurate results by observing the surrounding joints and local features of the image. In order to further dig out the dependence relationship of human body joints, we design a discriminator $D$ as shown in the bottom part of Figure~\ref{fig:framework}. The function of discriminator $D$ is to distinguish the right poses from the wrong poses, which do not satisfy the geometric relationship of human body. In contrast to previous work~\cite{chen2017adversarial}, we apply the gated graph neural network in discriminator $D$. As described in the section~\ref{word:GNN}, graph neural network can infer more reasonable joint information by using the internal dependence of human body joints.

Here, the discriminator $D$ takes the pose heatmap as input, which can be the results predicted from the generator or the ground-truth. The discriminator aims at predicting whether the pose is reasonable or not. Specifically, the output of $D$ is a $N \times 1$ vector, each value of which represents the rationality of corresponding body joint location in the input. Therefore, the output value of $D$ is in the range of $[$0$, $1$]$. The closer the value of the vector is to $1$, the more reasonable the body joint location is, and vice versa. 
The objective function for learning discriminator $D$ is defined as follows:
\begin{equation}
\begin{aligned}
\label{eq:bce_loss}
\quad L_{\mbox{D}} = \mathbb{E}[\mathrm{log} D(Y)] +\mathbb{E}[\mathrm{log} (1-D(X))],
\end{aligned}
\end{equation}
where $\mathbb{E}$ is the BCE loss. $D(X)$ and $D(Y)$ represent respectively the output of $D$ when given the predicted heatmaps $X$ and ground-truth heatmaps $Y$.

\subsection{Training Details}\label{}

This section will give the training process of our overall framework. Unlike some GAN based methods, we do not pretrain $G$ or $D$. After randomly initializing all the parameters, we train $G$ and $D$ alternatively following the general training process of GAN. Specifically, $G$ is trained for $3$ times then $D$ is trained for $1$ time. In the process of training $D$, we take the ground-truth heatmaps $Y$ as the input of $D$ and train $D$ to learn this is the reals. At the same time, we take the heatmaps $X$ generated by $G$ as the input of $D$ and train $D$ to learn this is the fakes. 
In the process of training $G$, $G$ is directly optimized to deceive $D$ through GAN. In other words, the $D$ will deem the heatmaps generated by current $G$ as the reals. 
Following the previous method~\cite{chen2017adversarial}, we mix the L2 loss of the generator with the BCE loss of the discriminator as the whole loss function. Therefore, the final objective function is defined as follows:
\begin{equation}
\label{eq:sum_loss}
\arg \min_{G} \max_{D} L_{G} + \alpha L_{D},
\end{equation}
where $\alpha $ is a hyper parameter to balance the L2 loss and the BCE loss. It is set to $0.01$ through experiments. Obviously, if $\alpha $ is equal to $0$, it means that the discriminator is not used to supervise the learning of $G$.

\section{Experiments}\label{}

\noindent{\bf Datasets.} 
To evaluate the proposed method, we conduct experiments on three public benchmark datasets for human pose estimation, \ie, extended Leeds Sports Poses (LSP)~\cite{johnson2010clustered}, MPII human pose~\cite{andriluka20142d} and COCO Keypoint Challenge~\cite{lin2014microsoft}.
The extended LSP dataset contains 11000 training images and 1000 testing images. For the MPII dataset, there are around 25000 images with 40000 annotated human poses, where 28000 human poses are used to train and 11000 human poses are used to test. 
Following the previous methods~\cite{chen2017adversarial,zhang2019human}, we also use the same setting to divide the MPII training poses into training set and validation set.
Our models are only trained on COCO train2017 dataset (includes $57000$ images and $150000$ person instances) with no extra data involved, validation is studied on the val2017 set(includes $5000$ images).

\noindent{\bf Implementation Details.} 
Our proposed model is implemented in Pytorch $1.0.0$. All experiments are run on a server with an Nvidia GEFORCE GTX 1080Ti GPU. 
We use Adam as the optimizer. For both benchmarks, the batch size is set to $64$ and the maximum epoch number is $140$. The learning rate is initialized to $0.001$ and reduced by $10$ times at $90$-th and $130$-th epochs. 

We crop the region of interest of each image based on the annotated person location from the dataset and warp it to the size of $256 \times 256$ for fair comparison with other methods. Data augmentation are adopted during training, such as random horizontal flipping, random rotation ($\pm 30$ degrees) and random scaling ($0.75$-$1.25$). Because the LSP contains less training data, we augment the training data of LSP with the MPII training images following \cite{insafutdinov2016deepercut,chen2017adversarial}.
The backbone network is ResNet-$50$ if not specially indicated in our experiments.

\noindent{\bf Evaluation Criteria.} 
We evaluate experimental results for the LSP dataset by using the Percentage Correct Keypoints (PCK) \cite{yang2012articulated}, which represents the percentage of correct detection. A joint location is believed to be correct if it falls within a normalized distance of the ground-truth. For the MPII dataset, we adopt the PCKh score \cite{andriluka20142d}, which uses a fraction of the head size to normalize the distance. 
For the COCO dataset, we use the official evaluation metric~\cite{lin2014microsoft} that reports the OKS-based AP (average precision) in the experiments, where the OKS (object keypoints similarity) is calculated from the distance between predicted joints and ground truth joints normalized by scale of the person.

\begin{table}
  \caption{The Comparison of PCK @0.2 on the LSP dataset.}
  \label{tab:lsp}
  \setlength{\tabcolsep}{10pt}{
  \resizebox{0.96\textwidth}{!}{
    \begin{tabular}{ccccccccc}
    \toprule
    Model&Head&Sho&Elb&Wri&Hip&Kne&Ank&Mean\\
    \midrule
    Belagiannis \etal \cite{belagiannis2017recurrent} & 95.2 & 89.0 & 81.5 & 77.0 & 83.7 & 87.0 & 82.8 & 85.2\\
    Lifshitz \etal \cite{lifshitz2016human} & 96.8 & 89.0 & 82.7 & 79.1 & 90.9 & 86.0 & 82.5 & 86.7\\
    Pishchulin \etal \cite{pishchulin2016deepcut} & 97.0 & 91.0 & 83.8 & 78.1 & 91.0 & 86.7 & 82.0 & 87.1\\
    Insafutdinov \etal \cite{insafutdinov2016deepercut} & 97.4 & 92.7 & 87.5 & 84.4 & 91.5 & 89.9 & 87.2 & 90.1\\
    Yang \etal \cite{yang2017learning} & 98.3 & 94.5 & 92.2 & 88.9 & 94.4 & 95.0 & 93.7 & 93.9\\
    Zhang \etal \cite{zhang2019human} & \bf 98.4 & 94.8 & 92.0 & 89.4 & 94.4 & 94.8 & 93.8 & 94.0\\
    \hline
    Our Model & 97.8 & \bf 95.5 & \bf 94.4 & \bf 92.9 & \bf 94.7 & \bf 95.6 & \bf 94.3 & \bf 95.1\\
    \bottomrule
    \end{tabular}}}
\end{table}

\subsection{Quantitative Results}
\label{}

\noindent{\bf Results on LSP.} 
Table~\ref{tab:lsp} lists the comparison between our method and previous methods, which are evaluated with PCK scores at a normalized distance of $0.2$. Following previous methods~\cite{yang2017learning}, the MPII training set is also added to the training set of extended LSP. From this table, we can see that our proposed approach outperforms all the previous methods. Compared with the previous best method~\cite{zhang2019human}, the performance improvement is $1.1\%$.
In particular, for some more challenging body parts such as wrist, ankle, we achieves $3.5\%$, and $0.5\%$ improvement compared to this method respectively.

\begin{table}[t!]
    \caption{The Comparison of PCKh @0.5 on the MPII validation set.}
    \label{tab:mpii}
    \setlength{\tabcolsep}{10pt}{
    \resizebox{0.96\textwidth}{!}{
    \begin{tabular}{cccccccccc}
    \toprule
    Model&Backbone&Head&Sho&Elb&Wri&Hip&Kne&Ank&Mean\\
    \midrule
    Hourglass~\cite{newell2016stacked} & $-$ & 96.0 & \bf 96.3 & 90.3 & 85.4 & 88.8 & 85.0 & 81.9 & 89.2\\
    SimpleBaseline~\cite{xiao2018simple} & ResNet-$152$ & 97.0 & 95.9 & 90.3 & 85.0 & 89.2 & 85.3 & 81.3 & 89.6\\
    LFP~\cite{yang2017learning} & $-$ & 96.8 & 96.0 & 90.4 & 86.0 & 89.5 & 85.2 & 82.3 & 89.6\\
    DLCM~\cite{tang2018deeply} & $-$ & 95.6 & 95.9 & \bf90.7 & \bf 86.5 & \bf 89.9 & 86.6 & 82.5 & 89.8\\ 
    AL~\cite{ryou2019anchor} & $-$ & 96.5 & 96.0 & 90.5 & 86.0 & 89.2 & \bf 86.8 & \bf 83.7 & \bf 89.9\\
    PGCN~\cite{bin2020structure} & ResNet-$50$ & $-$ & $-$ & $-$ & 83.6 & $-$ & $-$ & 80.8 & 88.9\\
    \hline
    Our Model & ResNet-$50$ & 96.4 & 95.3 & 89.1 & 84.2 & 89.1 & 84.8 & 80.4 & 89.0\\
    Our Model & ResNet-$152$ & \bf 97.1 & 95.9 & 90.4 & 85.1 & 89.1 & 85.8 & 81.5 & 89.8\\
    \bottomrule
    \end{tabular}}}
\end{table}

\noindent{\bf Results on MPII.} 
This subsection compares the performance of the MPII dataset between our method and some previous methods. 
The PCKh score at a normalized distance of $0.5$ is used to evaluate the performance. 
For simplicity, we directly use the PCKh scores in~\cite{sun2019deep}. Table~\ref{tab:mpii} shows the final results on MPII validation set. As shown in this table, our method obtains the PCKh score of $89.8\%$ on the MPII validation set, which is comparable to previous methods. 

\begin{table}[t!]
    \caption{Comparison with the 8-stage Hourglass~\cite{newell2016stacked}, CPN~\cite{chen2018cascaded}, SimpleBaseline~\cite{xiao2018simple}, ECSN~\cite{su2019multi} and PGCN~\cite{bin2020structure} on the COCO val2017 dataset. Their results are cited from~\cite{chen2018cascaded,xiao2018simple,su2019multi,bin2020structure}. OHKM means the model training with the Online Hard Keypoints Mining.}
    \label{tab:coco}
    \setlength{\tabcolsep}{20pt}{
    \resizebox{1\textwidth}{!}{
    \begin{tabular}{ccccc}
    \toprule
    Method&Backbone&Input Size&OHKM&AP\\
    \midrule
    $8$-stage Hourglass & $-$ & $256\times192$ & $\times$ & 66.9\\
    $8$-stage Hourglass & $-$ & $256\times256$ & $\times$ & 67.1\\
    CPN & ResNet-$50$ & $256\times192$ & $\times$ & 68.6\\
    CPN & ResNet-$50$ & $384\times288$ & $\times$ & 70.6\\
    CPN & ResNet-$50$ & $256\times192$ & $\checkmark$ & 69.4\\
    CPN & ResNet-$50$ & $384\times288$ & $\checkmark$ & 71.6\\
    SimpleBaseline & ResNet-$50$ & $256\times192$ & $\times$ & 70.4\\
    SimpleBaseline & ResNet-$50$ & $384\times288$ & $\times$ & 72.2\\
    ECSN & ResNet-$50$ & $256\times192$ & $\checkmark$ & 72.1\\
    ECSN & ResNet-$50$ & $384\times288$ & $\checkmark$ & 73.8\\
    PGCN & ResNet-$50$ & $256\times192$ & $\times$ & 71.1\\
    PGCN & ResNet-$50$ & $384\times288$ & $\times$ & 72.9\\
    \hline
    SimpleBaseline with CFN & ResNet-$50$ & $256\times192$ & $\times$ & 72.2\\
    Our overall method & ResNet-$50$ & $256\times192$ & $\times$ & 72.8\\
    SimpleBaseline with CFN & ResNet-$50$ & $384\times288$ & $\times$ & 73.8\\
    Our overall method & ResNet-$50$ & $384\times288$ & $\times$ & \bf 74.4\\
    \bottomrule
    \end{tabular}}}
\end{table}

\noindent{\bf Results on COCO.} 
Table~\ref{tab:coco} shows our result with the 8-stage Hourglass~\cite{newell2016stacked}, CPN~\cite{chen2018cascaded}, SimpleBaseline~\cite{xiao2018simple}, ECSN~\cite{su2019multi} and PGCN~\cite{bin2020structure} on COCO validation set. 
The human detection AP reported in Hourglass, CPN, ECSN and PGCN is $55.3$. The human detection AP of ours is the same $56.4$ as SimpleBaseline.
Compared with Hourglass~\cite{newell2016stacked}, both methods use an input size of $256\times192$ and no Online Hard Keypoints Mining(OHKM) involved, our method has an improvement of $5.9$ AP.
CPN~\cite{chen2018cascaded}, SimpleBaseline~\cite{xiao2018simple}, ECSN~\cite{su2019multi}, PGCN~\cite{bin2020structure} and our method use the same backbone of ResNet-$50$. When OHKM is not used, our method outperforms CPN by $4.2$ AP for the input size of $256\times192$ and $3.8$ AP for the input size of $384\times288$. Compared with SimpleBaseline, our method outperforms $2.4$ AP for the input size of $256\times192$ and $2.2$ AP for the input size of $384\times288$. 
When OHKM is used in CPN, our method outperforms CPN by $2.8$ AP for both input sizes. 
Compared with ECSN, our method outperforms $0.7$ AP for the input size of $256\times192$, and $0.6$ AP for the input size of $384\times288$.
Compared with PGCN, our method outperforms $1.7$ AP for the input size of $256\times192$, and $1.5$ AP for the input size of $384\times288$.
Therefore, we can conclude that our method has comparable results but is simpler.

\begin{figure}[t!]
\begin{center}
\includegraphics[width=0.92\linewidth]{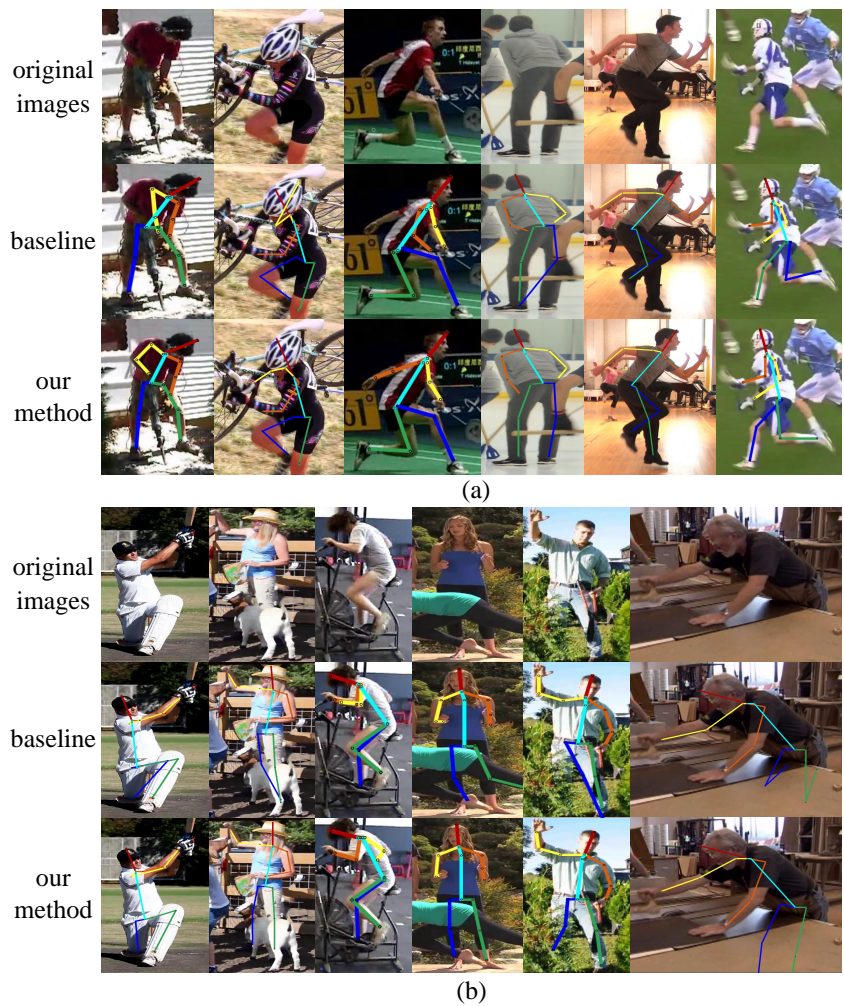}
\end{center}
\vspace{-8mm}
\caption{Sample estimation results on the MPII test set. 
The first row: original images. 
The second row: results by SimpleBaseline~\cite{xiao2018simple}. 
The third row: results by our overall method. 
(a) and (b) represent results for different poses.}
\label{fig:vis2}
\end{figure}

\begin{figure}[t!]
\begin{center}
\includegraphics[width=0.96\linewidth]{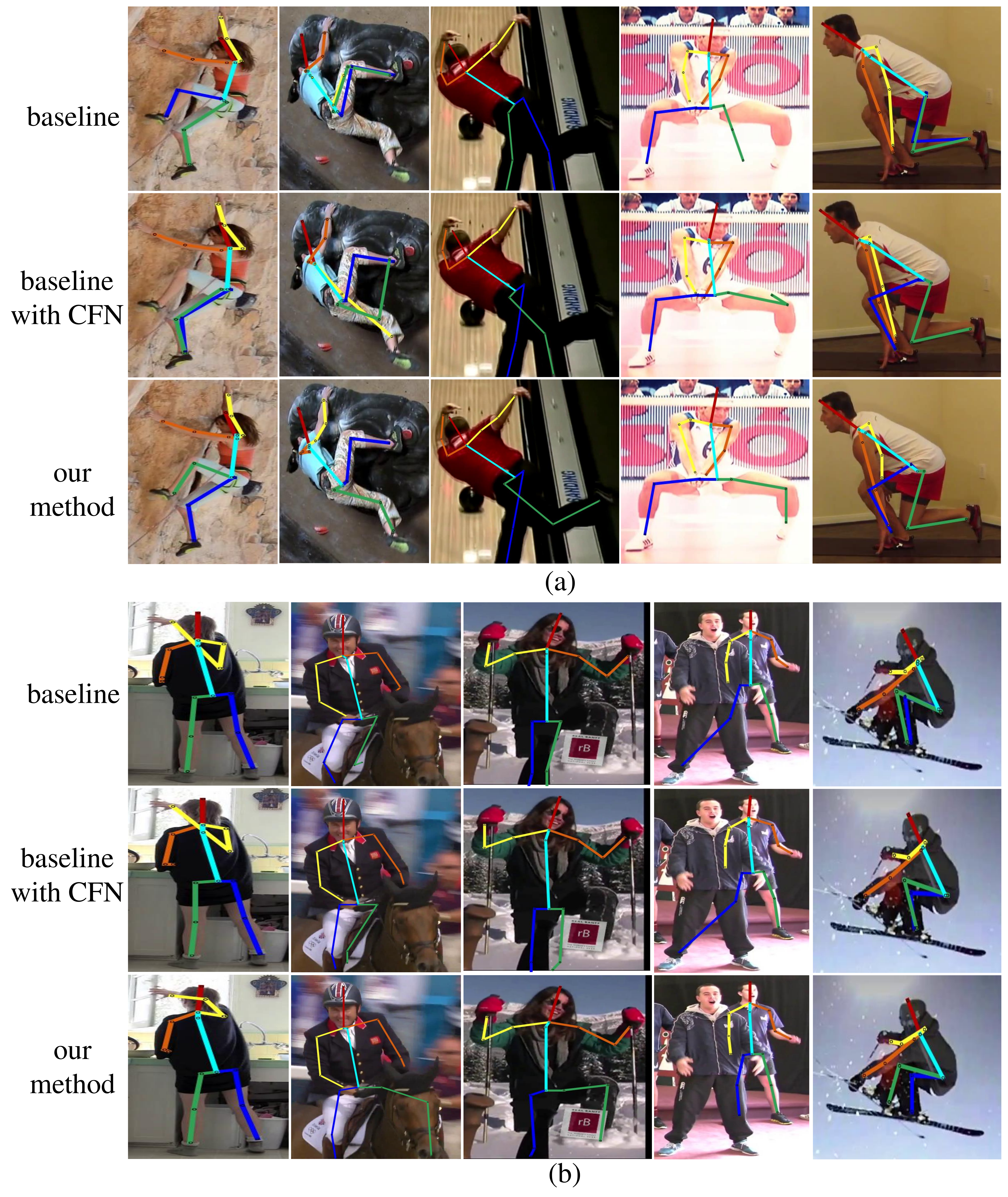}
\end{center}
\vspace{-5mm}
\caption{More examples show our approach on the MPII test set. 
The first row: results by SimpleBaseline~\cite{xiao2018simple}. 
The second row: results by SimpleBaseline with CFN. 
The third row: results by our overall method.}
\label{fig:vis3}
\end{figure}

\begin{figure}[t!]
\begin{center}
\includegraphics[width=0.96\linewidth]{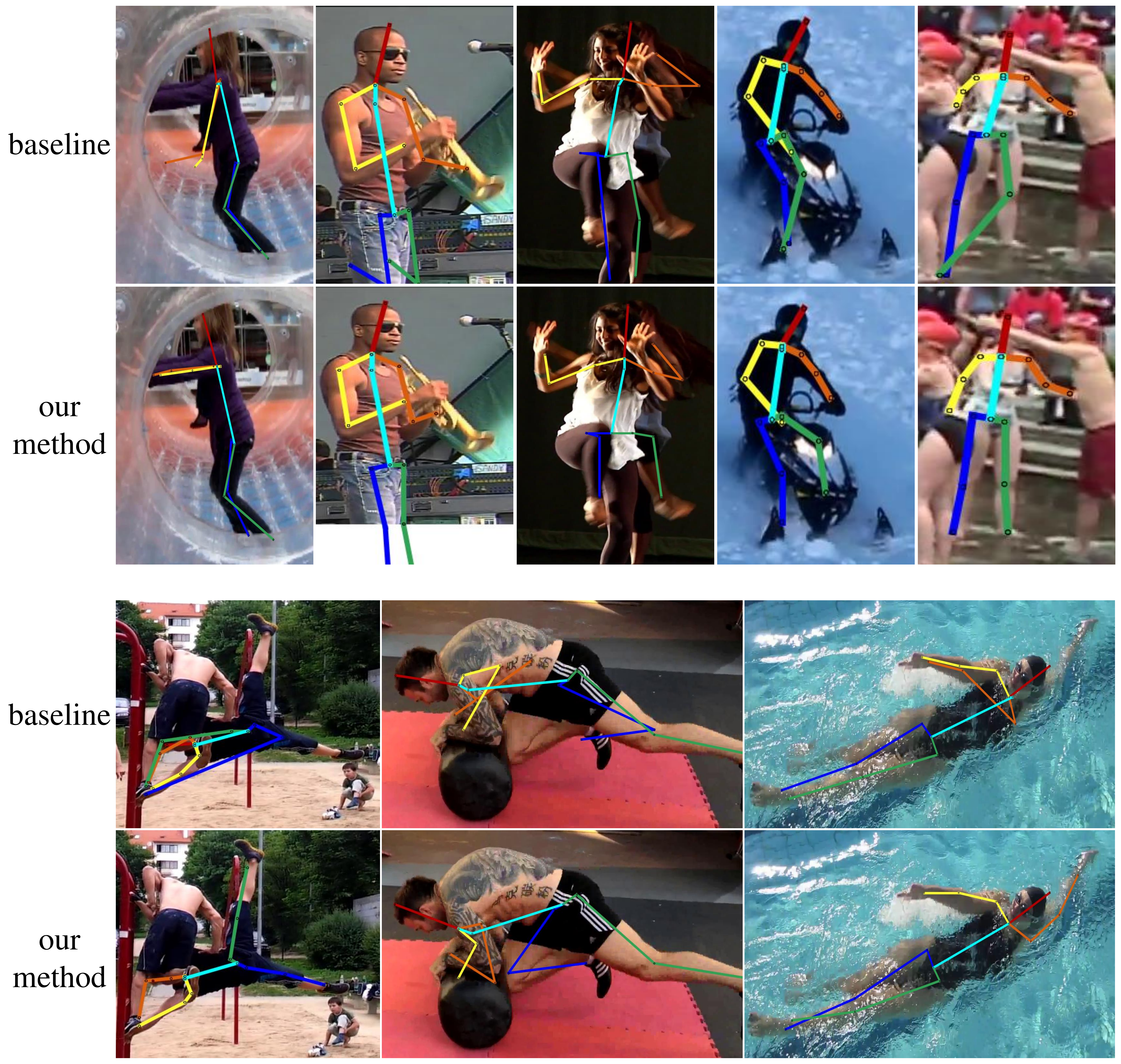}
\end{center}
\vspace{-5mm}
\caption{Failure examples on the MPII test set. 
The first row: results by SimpleBaseline~\cite{xiao2018simple}. 
The second row: results by our overall method.}
\label{fig:vis4}
\end{figure}

\subsection{Qualitative Comparisons}
\label{}

In this section, we give the qualitative comparison between our method and the SimpleBaseline~\cite{xiao2018simple}. Figure~\ref{fig:vis2} provides some examples of our model and SimpleBaseline on the MPII test set. 
We can see that it is difficult for the SimpleBaseline to generate reasonable body poses with large deformation and occlusion. However, our method gets more reasonable human pose. This may be because of the semantic information and the internal dependence of body joints learned in the adversarial training process. For example, the associated knee joints can help to infer the precise location of ankles in our model as shown in {col. 4-6} of Figure~\ref{fig:vis2}.

In (a) of Figure~\ref{fig:vis2}, SimpleBaseline produces some weird joints' location, especially for the joints on body limbs. Although SimpleBaseline has learned most of features about body joints, it may locate some body joints to the surrounding area rather than the right one without the joints dependence learned. 
In (b) of Figure~\ref{fig:vis2}, some body limbs are invisible due to self-occlusion and other-occlusion. In these examples, SimpleBaseline fails to produce reasonable poses. In contrast, our method successfully predicts the correct body position even under some difficult situations. This may be because of the graph structure introduced into the discriminator.

Figure~\ref{fig:vis3} shows more examples generated on the MPII test set.
In (a) of Figure~\ref{fig:vis3}, although adding CFN module to the baseline can reduce the number of the weird joints' location generated by the baseline, there are still some wrong joints' location.
And adding overall method to the baseline may locate some body joints to the right one with the joints dependence learned. 
In (b) of Figure~\ref{fig:vis3}, when some body limbs are invisible due to self-occlusion and other-occlusion, adding CFN to baseline can not solve the problem of the weird joints' location generated by baseline.
On the contrary, our method still solves this problem very well.

Figure~\ref{fig:vis4} shows some failure examples generated by our method on the MPII test set.
As shown in Figure~\ref{fig:vis4}, our method may fail in some very challenging examples with larger deformation, overlap and occlusion.
In some examples, human can not recognize the right pose at a glance. Even if our method fails in these cases, it gets a more reasonable poses than SimpleBaseline.

\begin{table}[t!]
    \caption{Effect of each component of our method on the MPII validation set.}
    \label{tab:ablation}
    \setlength{\tabcolsep}{10pt}{
    \resizebox{0.96\textwidth}{!}{
    \begin{tabular}{ccccc}
    \toprule
    Model&CFN&Discriminator&GSN &Mean PCKh\\
    \midrule
    \setlength{\arraycolsep}{20pt} SimpleBaseline~\cite{xiao2018simple} & $\times$ & $\times$ & $\times$ & 88.53\\
    SimpleBaseline~\cite{xiao2018simple} with CFN & $\checkmark$ & $\times$ & $\times$ & 88.74\\
    SimpleBaseline~\cite{xiao2018simple} with a Discriminator & $\times$ & $\checkmark$ & $\times$ & 88.64\\
    SimpleBaseline~\cite{xiao2018simple} with GSN & $\times$ & $\checkmark$ & $\checkmark$ & 88.80\\
    Our overall method & $\checkmark$ & $\checkmark$ & $\checkmark$ & \bf 89.02\\
    \bottomrule
    \end{tabular} } }
\end{table}

\subsection{Ablation Study}\label{}
To show the influence of each component of our method, we conduct ablation study. Since the ground-truth location of MPII test part is not available, we report the PCKh 0.5 score on the validation set of MPII dataset. In all experiments, the flip test is used.

\noindent{\bf Cascade Feature Network.} 
To evaluate the effect of CFN, we conduct two experiments using SimpleBaseline with and without a CFN. Both of the two models are trained only using the same L2 loss defined in Eq.~\eqref{eq:mse_loss}. In other words, there is no discriminators (\ie, no GAN). As shown in the first two rows of Table~\ref{tab:ablation}, the mean PCKh score on the MPII validation set is improved by $0.21\%$ compared to the SimpleBaseline model. We think this is caused by gathering the feature of previous layers of the ResNet-$50$ to the deconvolutional layers. This manifests that the cascade feature assists the network in better understanding of the poses.

\noindent{\bf Graph Structure Network.}
In this paper, we design a graph structure in the discriminator. So a question is the performance improvement is from GAN or the graph structure. To answer this question, we conduct an experiment only using a discriminator without graph structure. In detail, the discriminator consists of only two fully connected layers. Its PCKh score is shown in the third row of Table~\ref{tab:ablation}. From this table, we can find the score is increased by $0.11\%$ compared to the SimpleBaseline model. This shows the performance can be improved by adding a simple discriminator to the SimpleBaseline. Moreover, we also investigate the effect of discriminator including the GSN. As shown in the Table~\ref{tab:ablation}, the PCKh on MPII validation set is 88.80\%, which is respectively $0.27\%$  and $0.16\%$ higher than that of SimpleBaseline model and SimpleBaseline with a discriminator. This indicates that the discriminator with GSN is more helpful to promote the performance. In our mind, this is due to that the GSN can fully explore the internal geometric dependence of pose joints and reduce unreasonable pose estimation. 

\begin{figure}[t!]
\begin{center}
\includegraphics[width=0.96\linewidth]{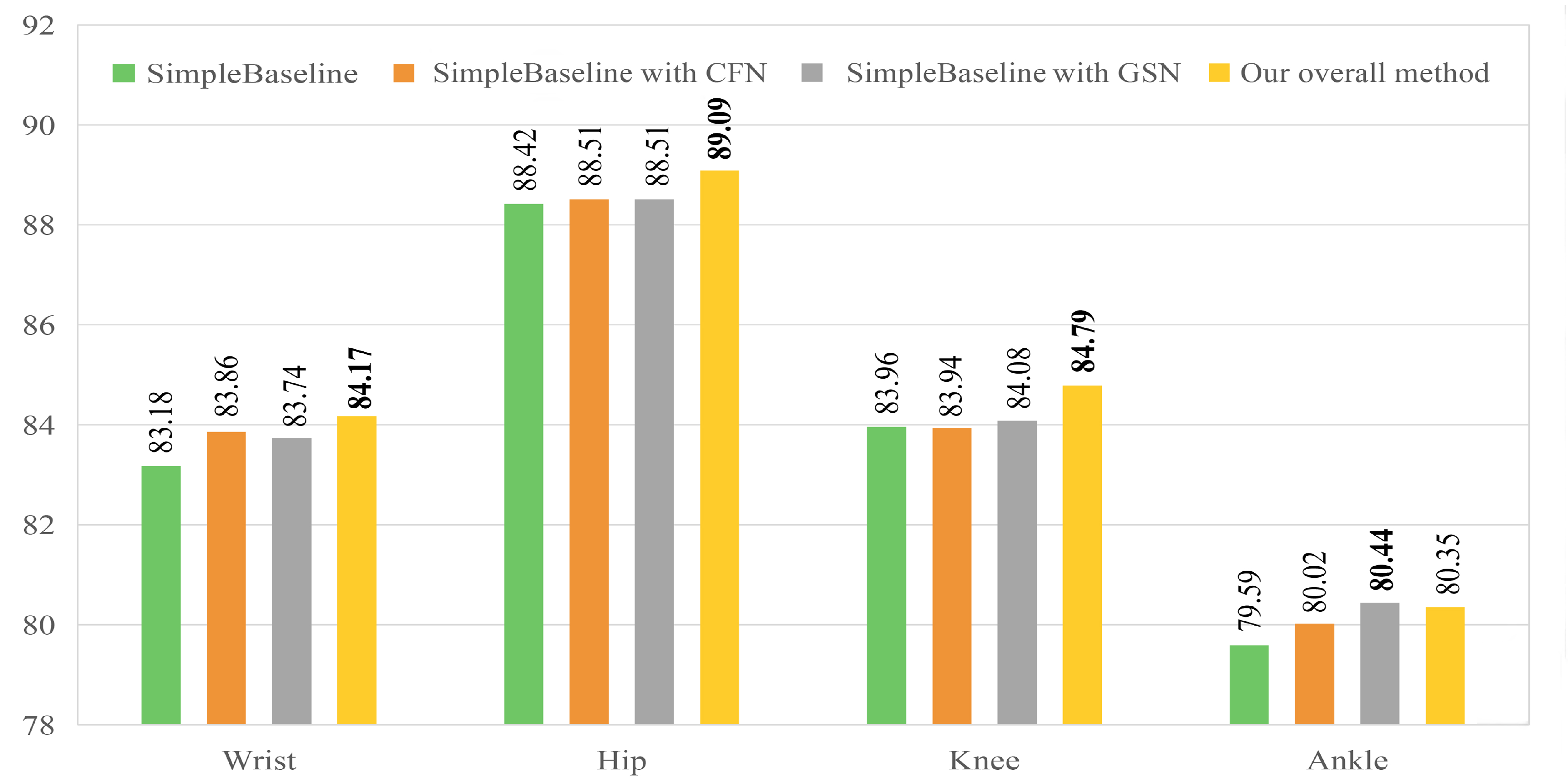}
\end{center}
\vspace{-7mm}
\caption{PCKh score of the wirst, hip, knee and ankle joints for different combination of components in our method.}
\label{fig:ablation}
\end{figure}

According to the above, adding CFN or a discriminator with GSN separately in the SimpleBaseline can improve the accuracy of human pose estimation. As shown in the last line of Table~\ref{tab:ablation}, using them simultaneously results in an increase of $0.5\%$, which is higher than only using one module.

Moreover, we also compare the PCKh score of wrist, hip, knee, ankle, whose localization difficulty is the largest in all human joints due to occlusion and higher freedom. As shown in Figure~\ref{fig:ablation}, for these four parts, our method achieves $0.99\%$, $0.67\%$, $0.83\%$ and $0.76\%$ improvement compared with the SimpleBaseline~\cite{xiao2018simple} respectively. This validates that the GSN capturing dependence of human joints can help to alleviate the occlusion problem. 

\begin{table}[t!]
    \caption{The compare the effect of the hyper parameter $\alpha $ on the performance on the MPII validation set.}
    \label{tab:alpha}
    \setlength{\tabcolsep}{35pt}{
    \resizebox{1\textwidth}{!}{
    \begin{tabular}{ccccc}
    \toprule
    $\alpha $ & $0.01$ & $0.1$ & $0.5$ & $1$\\
    \midrule
    Mean & 89.02 & 88.83 & 88.79 & 88.51\\
    \bottomrule
    \end{tabular}}}
\end{table}

We also compare the effect of the hyper parameter (\ie, variable $\alpha $) on the performance on the MPII validation set. 
As shown in Table~\ref{tab:alpha}, the performance reaches the best at $\alpha $ = $0.01$. It begins to degrade with the increase of $\alpha $. So, we set $\alpha$ to $0.01$.

\begin{table}[t!]
    \caption{The Analysis of the invisible human joints on the MPII validation set.}
    \label{tab:mpii_abla}
    \setlength{\tabcolsep}{10pt}{
    \resizebox{0.96\textwidth}{!}{
    \begin{tabular}{ccccc}
    \toprule
    Model & Num. of Invisible Joints & PCKh of Wrist & PCKh of Ankle & Mean\\
    \midrule
    \setlength{\arraycolsep}{10pt} 
    SimpleBaseline \cite{xiao2018simple} & $\geq$ 2 & 85.27 & 76.21 & 90.57\\
    Our overall method & $\geq$ 2 & \bf86.11 & \bf 77.67 & \bf90.92\\
    SimpleBaseline \cite{xiao2018simple} & $\geq$ 4 & 85.38 & 66.27 & 91.31\\
    Our overall method & $\geq$ 4 & \bf 87.11 & \bf72.62 & \bf 91.74\\
    \bottomrule
    \end{tabular}
    }
    }
\end{table}

\noindent{\bf Analysis of the Invisible Human Joints.}
In order to further show the effect of our method on the prediction of seriously occluded pose, we subdivide the validation set according to the number of invisible joints. Specifically, the number of samples with more than two invisible joints is 828, while the number of samples with more than four occluded joints is 489. 
Note that we do not train a new model, just compare the performance on these subsets of the MPII validation set. 
The results are shown in table~\ref{tab:mpii_abla}.
Although the location prediction of joints becomes more difficult with the increase of invisible joints, our method obtains better performance compared with the SimpleBaseline~\cite{xiao2018simple}. For the challenging body parts such as wrists and ankles, we achieve $0.84\%$ and $1.46\%$ improvement respectively compared to the SimpleBaseline when the number of invisible joints is bigger than $2$. Moreover, when the minimum number of invisible joints increases from $2$ to $4$, the performance gain becomes larger. These results indicate that our method can improve the pose estimation accuracy especially when some joints are occluded. 

\section{Conclusions}
\label{}
In this work, we propose a generative adversarial network consisting of CFN and GSN for human pose estimation. The CFN, which employs skip layers to combine features with different semantic information, is used as a generator to produce the location of human joints. Considering that the internal dependence correlation of human joints can be regarded as a natural graph structure, we develop a discriminator injected with a graph structure to effectively capture the dependence correlation of pose joints. Through adversarial learning, the generator can grasp the dependence of human joints, so the generator can produce pose estimation results with smaller error. Although we need to train the generator $G$ and the discriminator $D$, we only need the generator $G$ during testing. Therefore, the inference complexity of our method do not increase. To show the effectiveness of our model, we have done experiments and ablation studies on three widely used datasets, \ie LSP, MPII and COCO.

Currently, our graph is a tree-based structure, which may not completely capture the complex relationship of human joints, like the symmetric constraints. Therefore, we can design different graph structures, even dynamic structures to further improve the performance of human pose estimation. On the other hand, our model obtains better performance through injecting a graph structure. As a result, this model can be extended to other tasks including 3D human pose estimation, face landmark detection and semantic segmentation, where internal structure plays an important role.
\section*{Acknowledgements}
\label{}

This work was supported by the National Natural Science Foundation of China (No.61876152), the Ministry of Science and Technology Foundation funded project under Grant 2020AAA0106900, the National Natural Science Foundation of China (No.61902321, No.U19B2037), the China Postdoctoral Science Foundation funded project under Grant 2019M653746, and the Fundamental Research Funds for Central Universities of China under Grant 31020182019gx007. 

%% The Appendices part is started with the command \appendix;
%% appendix sections are then done as normal sections
%% \appendix

%% \section{}
%% \label{}

%% References: At least 5 are required 
%% If you have bibdatabase file and want bibtex to generate the
%% bibitems, please use
%%

%% else use the following coding to input the bibitems directly in the
%% TeX file.

%%\begin{thebibliography}{00}

%% \bibitem{label}
%% Text of bibliographic item

%%\bibitem{}

%\bibliographystyle{elsarticle-num} 
\bibliography{ref}

%%\end{thebibliography}

\end{document}